%% file: formatting-instructions-latex-2020.tex
\def\year{2020}\relax
\documentclass[letterpaper]{article} 
\usepackage{aaai20}  
\usepackage{times}  
\usepackage{helvet} 
\usepackage{courier}  
\usepackage[hyphens]{url}  
\usepackage{graphicx} 
\usepackage{graphicx}  
\usepackage{amsfonts}
\usepackage{amsmath}
\usepackage{amsthm}
\usepackage{algorithm}
\usepackage{algorithmic}
\usepackage{xspace}
\usepackage{graphicx}
\usepackage{subfigure}
\usepackage{wrapfig}
\usepackage{xcolor}
\usepackage{multirow}
\usepackage{cite}
\usepackage{natbib}

\title{A Quantitative Perspective on Values of Domain Knowledge for Machine Learning}

\author{\Large \textbf{Jianyi Yang} and \textbf{Shaolei Ren}\\
	University of California, Riverside
}
 
\begin{document}

\maketitle

\begin{abstract}
With the exploding popularity of machine learning, domain knowledge in various
forms has been playing a
crucial role in improving the learning performance, especially when training data is limited.
Nonetheless,  there is  little understanding of to what extent domain knowledge
can affect a machine learning task from a quantitative perspective.
To increase the transparency and rigorously explain the role of domain knowledge
in machine learning, we study the problem of quantifying the values
of domain knowledge in terms of its contribution to the learning performance
in the context of informed machine learning.
We propose a quantification  method based on Shapley value that fairly attributes
the overall learning performance improvement to different domain knowledge.
We also present Monte-Carlo sampling to approximate the fair value
of domain knowledge with a polynomial time complexity.
We run experiments of injecting symbolic domain knowledge
into  semi-supervised learning tasks 
on both MNIST and CIFAR10 datasets, providing
quantitative values of different symbolic knowledge
and rigorously explaining how it affects the machine learning performance
in terms of test accuracy.
\end{abstract}

\input{introduction}

\input{relatedwork}

\input{formulation}

\input{method}

\input{simulations}

\input{conclusion}

\clearpage

\bibliographystyle{plain}

\input{main.bbl}
\end{document}

%% file: introduction.tex
\section{Introduction}
Machine learning has achieved great success in numerous fields,
such as computer vision, natural language processing, security, and robotics.
The current success of machine learning, or more generally artificial intelligence (AI),
 owes much to the huge amount of available data.
 That being said, a priori domain knowledge, which comes from various sources and has multiple forms, also plays an increasingly more crucial role in the development of machine learning and may lead to new breakthroughs \cite{Informed_ML_von19,Generalizing_few_shot_wang2020,Improving_DL_Knowledge_Constriant_borghesi2020improving}.

Domain knowledge has already been playing a part in current machine learning pipelines in many real-world applications. Specifically,  machine learning with domain knowledge
injection (a.k.a informed machine learning \cite{Informed_ML_von19}) can be accomplished in different ways. For example, in computer vision, image pre-processing such as flipping \cite{One-shot_domain_translation_benaim2018one,Attentive_shyam2017attentive}, translation \cite{One-shot_domain_translation_benaim2018one}, cropping, \cite{Low_shot_imprinted_qi2018low}, scaling\cite{Fine_grained_meta_zhang2018fine} can help to augment data, which exploits the prior domain knowledge of image invariance. For another example, differential equations and logic rules from physical sciences and/or common knowledge provide additional constraints or new regularization terms for model training \cite{Interaction_learning_physics_battaglia2016interaction,Improving_DL_Knowledge_Constriant_borghesi2020improving,Injecting_knowledge_NN_silvestri2020injecting,Incorporating_domain_knowledge_muralidhar2018incorporating,Semantic_Loss_smbolic_xu2018semantic}. Moreover, knowledge graphs can provide relational knowledge for reasoning or feature representation in deep learning \cite{Symbolic_graph_CNN_liang2018symbolic,GRAM_choi2017gram}. 
In recent years, prior (domain) knowledge from one or more multiple teacher networks can also be injected into a learning task (i.e., student network) via distillation \cite{DNN_Knowledge_Distillation_Hinton_NIPS_DL_Workshop_2015_44873,DNN_Knowledge_Distillation_MultipleTeachers_KDD_2017_10.1145/3097983.3098135}.
As evidenced by many examples, such prior domain knowledge can effectively address the lack of labeled
data commonly encountered in machine learning tasks, improving the learning performance
in terms of various goals such as accuracy, generalization, and/or robustness.

In machine learning tasks, we  need not only a model with good performance, but also model explainability \cite{molnar2020interpretable,mythos_interpretability_lipton2018mythos}.
The model explainability can help us better understand the model, use the model with more trust, or more quickly debug the deployed model. There have been many prior studies
to address model explainability (especially in the context of deep neural
networks or DNNs) from various perspectives, such as model features \cite{molnar2020interpretable}, model inference \cite{Explaining_DNN_camburu2020explaining}, and contribution of individual data samples \cite{Data_Shapley_ghorbani2019data}.
Nonetheless, despite the increasingly wider usage of informed machine learning and
and the clear advantages over its \emph{uninformed}
counterpart, a quantitative understanding  of the actual contribution of
domain knowledge injected into machine learning tasks has been lacking.
This creates ambiguity about the role of domain knowledge in machine learning
and decreases the explainability of the resulting models.

A foundation of domain knowledge explainability is to quantify the value or contribution of domain knowledge for machine learning tasks.
In the first place, quantification of domain knowledge contribution gives us a better
transparency of informed machine learning, which is part
of the ``\emph{right
to explanation}'' legal requirement imposed
in Europe for
automated decision systems
 \cite{Goodman_Flaxman_2017}.
In addition, by domain knowledge quantification, we can know the actual contribution of different domain knowledge, which provides us
with a principled guideline about what kind of domain knowledge is mostly needed to be collected in similar machine learning tasks. Last but not least, domain knowledge injected in machine learning may be implicit, inaccurate, and/or even corrupted.
With domain knowledge quantification, we can make an informed decision as to which domain knowledge to utilize, since acquiring domain knowledge (e.g., coming from experts) may be quite expensive.

Nonetheless, it is non-trivial to quantify the value of domain knowledge in terms of
its contribution of the performance of a learning  task.
First, domain knowledge has various forms and can be injected in machine learning in different ways \cite{Informed_ML_von19}. Moreover, different domain knowledge is combined together and injected into a machine learning task, and the overall performance is typically not a simple linear combination of the performance corresponding to each piece of domain knowledge. Last but not least, we need to guarantee \emph{fairness}
for the quantification of domain knowledge in terms of its contribution
to model performance.

To address these challenges, we leverage a classic tool from game theory --- Shapley value
--- which 
provides a fair distribution of total payoff among several participants \cite{shapley1953value}.
Shapley value has been recently used in machine learning model explainability to quantify the contribution of features \cite{Explaining_feature_vstrumbelj2014explaining}, training data samples \cite{Data_Shapley_ghorbani2019data}, algorithms \cite{Jointly_algorithm_data_quantification_yona2019s}, and neurons  \cite{Neuron_Shapley_ghorbani2020neuron}.
But, to our knowledge, quantifying the contribution of domain knowledge
for machine learning is a novel perspective on model explainability
that has not been considered. It will only become more important
as informed machine learning becomes increasingly ubiquitous.

Our main contributions are summarized as follows.
\begin{itemize}
	\item We formulate the problem of quantifying the value of domain knowledge in informed machine learning, with a focus on the common approach
to using regularization for domain knowledge injection into machine learning tasks.
	\item We present a quantification method based on Shapley value to measure the value of domain knowledge in a fair manner. Additionally, we address the computational cost
of Shapley value and give an approximated algorithm based on Monte-Carlo sampling, which achieves a polynomial time complexity.
	\item As examples of quantifying the value of domain knowledge, we perform experiments on image classification
tasks using semi-supervised learning over datasets of MNIST \cite{MNIST} and CIFAR10 \cite{CIFAR}. Specifically, we consider symbolic domain knowledge and inject it into a learning task by adding a knowledge-related loss to the learning objective. Our experiments show the performance improvement by injecting different combinations of domain knowledge, from which we calculate the value of each domain knowledge for quantitative explainability.
Our results also quantify and explain the performance degradation due to imperfect domain knowledge (on MNIST).
\end{itemize}

%% file: relatedwork.tex
\section{Related Work}

\textbf{Explanation of DNN models.}
Since most of the current DNN models are black-box and hard to explain, model interpretability/explainability has attracted much attention from both academia and industry \cite{molnar2020interpretable,mythos_interpretability_lipton2018mythos}. The techniques to explain DNN or AI models can be broadly categorized into intrinsic interpretation techniques and post-hoc interpretation \cite{molnar2020interpretable}, depending on whether the adopted explainability method is integrated with the learning process and affects the model. Intrinsic explainability techniques usually rely on some explainable/interpretable models like regression models, decision tree \cite{Interpreting_decision_tree_zhang2019interpreting},  attention models \cite{Uncertainty_attention_interpretation_heo2018uncertainty}, etc. More recently,  intrinsic explainability techniques are also used in deep learning to achieve self-explanatory neural networks \cite{Robust_interpretability_selfexplaining_NN_alvarez2018towards,Interpretable_deep_CNN_shen2019interpretable}. Intrinsic explainability can have more accurate explanations for the output of DNN models, but it may also decrease the learning performance to some extent.

Unlike intrinsic explainability, post-hoc explainability
uses a
 new explainable model to
 explain another more complex model after the model is
trained.
 Many model-based methods have been proposed for post-hoc interpretability/explainability  \cite{Simple_model_based_importance_measure_greenwell2018simple, Anchors_model_agnostic_ribeiro2018anchors,Model_agnostic_intertability_ribeiro2016model}.
By contrast,  explainability based on Shapley value is a model-free method and has been becoming increasingly popular for AI explanation due to its flexibility. Shapley value is proved to be the only method that satisfies a set of axioms of fairness in payoff distribution problems \cite{shapley1953value}.
Among existing  techniques based on Shapley value, \cite{Explaining_feature_vstrumbelj2014explaining} explains the features of model input,  \cite{Data_Shapley_ghorbani2019data} studies explanation of the value of data samples, and \cite{Neuron_Shapley_ghorbani2020neuron} quantifies the contribution of neurons in neural networks and proposes a Bayesian optimization algorithm to reduce the evaluation complexity. The joint contribution of learning algorithms and dataset are studied in \cite{Jointly_algorithm_data_quantification_yona2019s} using Shapley value.  Also, \cite{Explaining_DNN_Polynomial_Shapleyancona2019explaining} propose a polynomial algorithm to compute Shapley values and explain input features for DNNs.
In this paper, we leverage Shapley value for model explainability
from a new perspective ---
quantifying the value of domain knowledge.

\textbf{Informed Machine Learning}
Many techniques have been developed to inject domain knowledge in machine learning tasks \cite{Informed_ML_von19,Generalizing_few_shot_wang2020}.
 One naive approach is to generate additional synthetic training data based on domain knowledge.
 For example,  samples generated from simulations or demonstration can help mitigate the data shortage problem in reinforcement learning \cite{Simulation_ML_deist2019simulation,RL_imperfect_demonstraion_gao2018reinforcement,DQN_demonstration_hester2017deep,Physics_pgnn_karpatne2017physics,Using_Simulation_BO_robots_rai2019using}.  Additionally, data can be generated by knowledge graphs \cite{Hybrid_knowledge_objective_detection_jiang2018hybrid}, generative models
 \cite{Low_shot_augmentation_gao2018low}, etc.
Domain knowledge can also benefit machine learning tasks by choosing hypothesis sets like regression model and neural architectures. A good hypothesis set can improve the learning performance as well as reduce the training complexity. Examples include domain knowledge based neural architecture search \cite{ModuleNet_chen2020modulenet,Knowledge_based_ANN_towell1994knowledge} and hyper-parameter tuning \cite{Towards_automatically_tuning_mendoza2016towards}.

Injecting domain knowledge into machine learning tasks via a loss term in
the training objective function is another common method. For example, AI models applied in real world are often trained on the optimization objective with constraints originated from domain knowledge \cite{Improving_DL_Knowledge_Constriant_borghesi2020improving,Injecting_knowledge_NN_silvestri2020injecting,Incorporating_domain_knowledge_muralidhar2018incorporating}. Knowledge distillation from
one or more multiple teacher networks also falls into this category of knowledge injection \cite{DNN_Knowledge_Distillation_Hinton_NIPS_DL_Workshop_2015_44873,DNN_Knowledge_Distillation_MultipleTeachers_KDD_2017_10.1145/3097983.3098135}.
Moreover, recent studies have shown that logical rules can also be integrated into machine learning by including a domain knowledge related loss term \cite{Semantic_Loss_smbolic_xu2018semantic,Harnessing_DNN_logic_hu2016harnessing}.

To our knowledge, however, the values of  domain knowledge injected
into informed machine learning have not been rigorously or fairly quantified.

%% file: formulation.tex
\section{Informed Machine Learning}
A key goal of machine learning is to find the intrinsic hypothesis for a learning task. Typically, a hypothesis $h$ represents a mapping from an input variable $X$ to an output variable $Y$. In some cases, a hypothesis is a distribution over the mapping.  In
conventional machine learning, a training dataset $\mathcal{D}$ is provided and a hypothesis set  $\mathcal{H}$ is determined, usually by choosing a learning model $h(X,\theta)$. With a certain loss function and learning algorithm, the model $h(X,\theta)$ is learnt to approximate the optimal hypothesis $h^*=h(X,\theta^*)$  which has the optimal performance among the chosen hypothesis set $\mathcal{H}$. For example, given
a performance metric $R(h)$ (which can be risk, cost, or negative reward/utility), the optimal hypothesis in the chosen hypothesis set is $h^*=\arg\min_{h\in\mathcal{H}}R(h)$.  By training based on the dataset $\mathcal{D}$, we get a final hypothesis $\hat{h}_{\mathcal{D}}=h(X,\hat{\theta}_{\mathcal{D}})$.

For a machine learning task, various forms of domain knowledge from multiple sources can
be integrated into the machine learning pipeline for improving the model performance. For example, domain knowledge can be used for data augmentation \cite{Informed_ML_von19,Generalizing_few_shot_wang2020}
by generating synthetic data samples via simulations
\cite{Simulation_ML_deist2019simulation,DQN_demonstration_hester2017deep,Physics_pgnn_karpatne2017physics},
data transforming \cite{Low_shot_imprinted_qi2018low,Fine_grained_meta_zhang2018fine}, and generative models \cite{GAN_goodfellow2014generative, Low_shot_augmentation_gao2018low}.
With more (independent) samples in the augmented dataset $[\mathcal{D}, \tilde{\mathcal{D}}_K]$, the generalization error can be reduced \cite{Aprimer_PAC_guedj2019primer}.
Also, domain knowledge is important for choosing a hypothesis set $\mathcal{H}$. This is usually done by choosing a learning model $h_K(X,\theta)$, such as knowledge-based neural architecture search \cite{ModuleNet_chen2020modulenet} and hyper-parameter tuning \cite{Towards_automatically_tuning_mendoza2016towards}.
Moreover, domain knowledge can be used to decide whether some final  hypothesis should be discarded \cite{Informed_ML_von19}.

In this paper, we focus on another
important approach
to domain knowledge injection: 
adding knowledge-based constraints or  loss terms
 as part of the objective function
 \cite{Improving_DL_Knowledge_Constriant_borghesi2020improving,Injecting_knowledge_NN_silvestri2020injecting,Incorporating_domain_knowledge_muralidhar2018incorporating,Semantic_Loss_smbolic_xu2018semantic}.
Specifically, with a data-based loss $L(h, \mathcal{D})$ used in conventional machine learning and a loss $L_{K_n}(h, \bar{\mathcal{D}}_{n})$ based on domain knowledge $K_n$ 
applied to its corresponding (possibly unlabelled) dataset $\bar{\mathcal{D}}_{n}$,
for $n=1,\cdots, N$,
the training objective based on domain knowledge can be expressed as
\begin{equation}\label{eqn:formulation}
\hat{h}=\arg\min_{h\in\mathcal{H}} L(h, \mathcal{D})+\sum_{n=1}^{N}\lambda_n L_{K_n}(h, \bar{\mathcal{D}}_{n}),
\end{equation}
where $\lambda_n\geq0$ is used to balance the pure data-based loss and knowledge-based loss. In addition to
knowledge-based constraints or losses,
 this formulation also applies to  data
 augmentation based on domain knowledge: $\bar{\mathcal{D}}_n$ is
 the synthetic training dataset generated based on domain knowledge
 $K_n$ and we have $L_{K_n}(h, \bar{\mathcal{D}}_{n})=L(h, \bar{\mathcal{D}}_n)$
 for $n=1,\cdots, N$.

Assume that $R(h)$  is the performance metric of interest in a machine learning task given
a set containing $N$ pieces of domain knowledge $\mathcal{K}=\left\lbrace K_1, K_2,\cdots, K_N\right\rbrace$. 
For conciseness, $K_n$ represents the $n$-th domain knowledge as well as the corresponding
(possibly unlabelled) dataset $\bar{\mathcal{D}}_{n}$.
With the final hypothesis $\hat{h}_{\mathcal{D},\mathcal{K}}$ learnt based on both training dataset $\mathcal{D}$ and domain knowledge in $\mathcal{K}$, the learning performance is denoted as $R(\hat{h}_{\mathcal{D},\mathcal{K}})$.

%% file: method.tex
\section{The Value of Domain Knowledge}

 For informed machine learning with domain knowledge, the model performance
 highly relies on the included domain knowledge information.
 Naturally, different domain knowledge can have different contributions to the learning performance, and the performance metric can change a lot by adding, removing or revising a subset of the domain knowledge.
  Thus,  to explain the learnt model and achieve better transparency,
  it is crucial to quantify the value of each domain knowledge for
   the considered learning task in terms of the metric of interest.  Moreover,
   the quantification of domain knowledge values should be \emph{fair}.
   In this section,  we give a set of axioms for fair quantification of domain knowledge
   values and propose a method based on Shapley value to achieve fair quantification.

\subsection{Axioms for Fairness}

If a set of knowledge $\mathcal{K}=\left\lbrace K_1, K_2,\cdots, K_N\right\rbrace$, together with a labelled training dataset $\mathcal{D}$, jointly determines the model performance,
the performance improvement contributed by the domain knowledge set $\mathcal{K}$ is denoted as
\begin{equation}
V(\mathcal{K})=R(\hat{h}_{\mathcal{D}})-R(\hat{h}_{\mathcal{D},\mathcal{K}}),
\end{equation}
where $R(\hat{h}_{\mathcal{D}})$ is the performance without  domain knowledge set $\mathcal{K}$. 
Alternatively, if the domain  knowledge is already
included into a machine learning task, the performance improvement due to the specific domain knowledge set $\mathcal{K}$ can be expressed as
\begin{equation}
V(\mathcal{K})=R(\mathbb{E}_{\mathcal{K}}[\hat{h}_{\mathcal{D},\mathcal{K}}])-R(\hat{h}_{\mathcal{D},\mathcal{K}}),
\end{equation}
where $\mathbb{E}_{\mathcal{K}}[\hat{h}_{\mathcal{D},\mathcal{K}}]$ is the hypothesis averaged over all the possible domain knowledge. 

To quantify the contribution of individual domain knowledge,
the performance improvement $V(\mathcal{K})$ should be attributed
into individual contributions $\phi(K_n),n=1,2,\cdots,N$ by different domain knowledge in the set $\mathcal{K}$. Importantly, a \emph{fair} attribution should meet the following basic axioms \cite{shapley1953value,Data_Shapley_ghorbani2019data}.\\
\textbf{Axiom 1} (Efficiency)
The performance improvement contributed by the entire
domain knowledge set $\mathcal{K}$ is the sum of contributions by
individual domain knowledge, i.e.
\begin{equation}
V(\mathcal{K})=\sum_{n=1}^{N}\phi(K_n).
\end{equation}\\
\textbf{Axiom 2} (Symmetry)
If two different domain knowledge $K_n$ and $K_m$ are exchangeable and not distinguishable in terms of the performance improvement, they should have equal contributions. Thus, for any  knowledge subset $\mathcal{K}'\subset \mathcal{K}\setminus\left\lbrace K_n,K_m\right\rbrace $, we have
\begin{equation}
V(\mathcal{K}'\cup\left\lbrace K_n\right\rbrace )= V(\mathcal{K}'\cup\left\lbrace K_m\right\rbrace )= \gg \phi(K_n)=\phi(K_m).
\end{equation}\\
\textbf{Axiom 3} (Null-player)
If individual domain knowledge  $K_n$ in $\mathcal{K}$ has no contribution to the performance improvement $V(\mathcal{K})$, the attributed contribution of $K_n$ is zero. That is, for any domain knowledge subset $\mathcal{K}'\in \mathcal{K}\setminus\left\lbrace K_n\right\rbrace $, we have
\begin{equation}
V(\mathcal{K}'\cup\left\lbrace K_n\right\rbrace )=V(\mathcal{K}' )= \gg \phi(K_n)=0.
\end{equation}

\subsection{Domain Knowledge Shapley}

Since domain knowledge in the set $\mathcal{K}$ combined together achieves performance improvement $V(\mathcal{K})$, it is non-trivial to find a contribution attribution method that satisfies all the above three axioms. To address this,
we provide a attribution method based on Shapley value --- Domain Knowledge Shapley
---
to quantify the contribution of different domain knowledge.
Importantly, Shapley value, a classic tool from
cooperative game theory, is proved to be the only method that satisfies the axioms (along
with other properties) for fair attribution, as will be introduced in detail next \cite{shapley1953value}.

In our problem, the performance improvement resulting from the
domain knowledge  set $\mathcal{K}$ can be seen as the total payoff to be distributed and each domain knowledge $K_n$ in the set $\mathcal{K}$ can be seen as a player in a cooperative game.
Thus, domain knowledge contribution is also the payoff distributed to each player.
 Shapley value with respect to the $n$-th player representing domain knowledge $K_n$ is calculated as
\begin{equation}\label{eqn:knowledgeShapley}
\phi(K_n)=\frac{1}{N}\sum_{\mathcal{K}'\subseteq \mathcal{K}\setminus\left\lbrace K_n \right\rbrace }\frac{V(\mathcal{K}'\cup \left\lbrace K_n \right\rbrace)-V(\mathcal{K}')}{\left ( \begin{matrix}
N-1	\\ 	|\mathcal{K}'|
	\end{matrix} \right )},
\end{equation}
where $\left ( \begin{matrix}
N-1	\\ 	|\mathcal{K}'|
\end{matrix} \right )$ is the number of combinations with size $|\mathcal{K}|$ from the total
knowledge set $\mathcal{K}$, and $V(\mathcal{K}'\cup \left\lbrace K_n \right\rbrace)-V(\mathcal{K}')$ is the marginal contribution of domain knowledge $K_n$ on the basis of a domain knowledge subset $\mathcal{K}'$.
Intuitively, Shapley value quantifies the contribution of domain knowledge $K_n$ as the average marginal contribution of $K_n$
over all the possible $\mathcal{K}'\subseteq \mathcal{K}\setminus\lbrace K_n\rbrace $.
The reason to average over the domain knowledge subset $\mathcal{K}'$
is \emph{fairness}: the order of adding each domain knowledge
for quantifying its marginal contribution should be equal for all domain knowledge.

\subsection{Approximate Method}

Although Shapley value is a provably fair in quantifying
the contribution of domain knowledge, it needs $2^{|\mathcal{K}|}$ evaluations in total,
accounting for different combinations of domain knowledge \cite{shapley1953value}.
Thus, the computational cost of Shapley value is
exponentially increases with the size of the knowledge set $|\mathcal{K}|$,
and can quickly become intolerable for some practical machine learning tasks
that have both high training and testing costs.
As a result, explaining DNN models (e.g., the contribution of training data
samples \cite{Data_Shapley_ghorbani2019data}) based on Shapley value
typically leverage approximate methods.

Monte-Carlo sampling is a common method to approximate the computation of Shapley
values in the context of DNN explainability \cite{Polynomial_Calculation_Shapley_castro2009polynomial,Bounding_error_Sampling_Shapley_maleki2013bounding}. Specifically, we denote $\varpi(\mathcal{K})$ as the set of permutation of $\mathcal{K}$ and rewrite domain knowledge Shapley in Eqn.~\eqref{eqn:knowledgeShapley} based
on \cite{Polynomial_Calculation_Shapley_castro2009polynomial} as follows:
\begin{equation}\label{eqn:shapley_permutation}
\begin{split}
\phi(K_n)\!=\!\sum_{O\in\varpi(\mathcal{K})}\frac{1}{ N!}
\left[ V(O_{\mathrm{Pre},n}\cup \left\lbrace K_n \right\rbrace)-V(O_{\mathrm{Pre},n})\right]
\end{split}
\end{equation}
where $O_{\mathrm{Pre},n}$ is the set of predecessors of $K_n$ in a permutation $O$.  Based on Eqn. \eqref{eqn:shapley_permutation}, the domain knowledge Shapley for $K_n$ can be estimated by sampling from the permutation of domain knowledge set $\mathcal{K}$. The approximated knowledge Shapley algorithm is given in Algorithm~\ref{alg:approx_know_shapley}.
Critically,
Shapley approximation based on Monte-Carlo sampling has a polynomial-time complexity
and has been proven to have a bounded error with a high probability \cite{Bounding_error_Sampling_Shapley_maleki2013bounding}.
\begin{algorithm}[!t]
	\caption{Approximated Domain Knowledge Shapley}\label{alg:approx_know_shapley}
	\begin{algorithmic}
		\REQUIRE  Domain knowledge set $\mathcal{K}$, performance evaluation $V$
		\ENSURE   Estimated domain knowledge Shapley $\hat{\phi}(K_n)$, $n=1,\cdots, N$
		\STATE\textbf{Initialization} $count=0$, $\hat{\phi}(K_n)=0, n=1,\cdots, N$
		\WHILE {$count<\mathrm{MaxIterNum}$}
		\STATE Randomly generate a permutation $O$ of $\mathcal{K}$
		\FOR {$n=1,\cdots, N$}
		\STATE   $O_{\mathrm{Pre},n}$: set of predecessors of $K_n$.
		\STATE   $\hat{\phi}(K_n)\!\!=\!\!\hat{\phi}(K_n)+\left[ V(O_{\mathrm{Pre},n}\cup \left\lbrace K_n \right\rbrace)-V(O_{\mathrm{Pre},n})\right]$
		\ENDFOR
		\STATE count=count+1
		\ENDWHILE
		\STATE $\hat{\phi}(K_n)=\hat{\phi}(K_n)/\mathrm{MaxIterNum}$, for $n=1,\cdots, N$.
	\end{algorithmic}
\end{algorithm}

While Monte-Carlo sampling can effectively reduce the
complexity, the evaluation for each marginal contribution
$V(O_{\mathrm{Pre},n}\cup \left\lbrace K_n \right\rbrace)-V(O_{\mathrm{Pre},n})$
can still be costly due to the requirement of calculating the values
of $V(O_{\mathrm{Pre},n}\cup \left\lbrace K_n \right\rbrace)$
and $V(O_{\mathrm{Pre},n})$. To further reduce
the cost of evaluating the learning performance with different
combinations
of domain knowledge, we can train the model over a smaller number
of epochs and also use a smaller dataset for testing.
Moreover, if the size of domain knowledge set is very large,
we can also train an predictor for the learning performance
that takes the domain knowledge subset
as input and directly estimates the resulting learning performance without training.
Such performance predictors are also used for avoiding
the cost of model training for each architecture
candidate and speeding up neural architecture search  \cite{DNN_NAS_ChamNet_Prediction_CVPR_2019_dai2019chamnet,DNN_NAS_APQ_JointSearch_ArchitecturePruningQuantization_SongHan_CVPR_2020_Wang_2020_CVPR}.

%% file: simulations.tex
\section{Experiments}

In this section, we run experiments
to quantify  the values of domain knowledge for two image classification tasks
under a semi-supervised setting for which domain knowledge can significantly improve
the learning performance.

\subsection{DNN with Symbolic Domain Knowledge}

In our experiments, the
domain knowledge comes from the symbolic label knowledge which can be expressed as a logical sentence \cite{Semantic_Loss_smbolic_xu2018semantic,Semantic_regularization_diligenti2017semantic}.
Specifically, a logical sentence is a set of assertions combined by relational expressions such as $\wedge$ (and), $\vee$ (or) and $\neg$ (not). For example, for states $S_1, S_2, S_3$, a logical sentence can be $\alpha=S_1 \vee (S_2 \wedge \neg S_3)$.  Assume that for a logical sentence $\alpha$ and a state $S$, $S\models \alpha$ means that the state $S$ satisfies the logical knowledge $\alpha$, or in other words, the logical sentence $\alpha$ can be written as $\alpha=S\vee\cdots$. In the previous example, we have $S_1\models \alpha$ and $(S_2 \vee \neg S_3)\models \alpha$.

A common approach to injecting symbolic knowledge in DNNs is adding a knowledge-based loss in the training objective \cite{Semantic_Loss_smbolic_xu2018semantic}.
 Consider a classification task
 where $X\in\mathcal{X}$ is the input  and $Y\in \mathcal{Y}=\left\lbrace1,2, \cdots, J\right\rbrace $ is the label.
We use
 a neural network $h(X,\theta)$ as our classification model. Denote the output of $h(X,\theta)$ as $\mathbf{\hat{P}}(X)=\left[\hat{P}_1(X),\hat{P}_2(X),\cdots,\hat{P}_J(X) \right] $ which is usually interpreted as the estimated probability that the corresponding label is the correct label. We include an additional loss term
  based on the symbolic knowledge as  proposed in \cite{Semantic_Loss_smbolic_xu2018semantic}, which is the negative logarithm of the estimated probability that the knowledge is satisfied. If $S$ represents a state of the sparse label space $\left\lbrace 0,1\right\rbrace^J $, given an domain knowledge $K$ in the logical sentence form and an input $X$,  the knowledge-based loss can be  expressed as
\begin{equation}\label{loss_onehot}
L_{K}(X)=-\log\sum_{S\models K}\prod_{j: S\models (Y=j)}\hat{P}_j(X) \!\!\!\!\prod_{j: S\models ( Y\neq j)}(1-\hat{P}_j(X)).
\end{equation}
For example, if the domain knowledge $K$ means that only one label in $\mathcal{Y}$ is the correct output, then the knowledge-based loss is
\begin{equation}\label{loss_know}
L_{K}(X)=-\log\sum_{j\in\mathcal{Y}}\hat{P}_j(X)\prod_{i\neq j}(1-\hat{P}_{i}(X)).
\end{equation}
Similarly, if the knowledge $K$ means that the correct label with respect to input $X$ is one of the labels in the set $\mathcal{Y}'\subseteq\mathcal{Y}$, the knowledge-based loss can be expressed as
\begin{equation}
L_{K}(X)=-\log\sum_{j\in\mathcal{Y}'}\hat{P}_j(X)\prod_{i\neq j}(1-\hat{P}_{i}(X)).
\end{equation}

The above domain knowledge injection method can be used to improve the learning performance
for semi-supervised classification tasks.
Specifically, we are given a training dataset $\mathcal{D}$ with input-label pairs
as well as an unlabelled dataset  $\bar{\mathcal{D}}_X$. Denote $L_{\mathrm{ce}}(x,y)$ as the cross-entropy loss of an instance $(x,y)$. The semi-supervised training objective with a set of knowledge $\mathcal{K}=\left\lbrace K_1,\cdots, K_N\right\rbrace $ injected is expressed as
\begin{equation}
L_{\mathrm{semi}}=\sum_{(x,y)\in \mathcal{D}}L_{\mathrm{ce}}(x,y)+\sum_{n=1}^N\lambda_n \sum_{x\in \bar{\mathcal{D}}_X}L_{K_n}(x).
\end{equation}

Given a test input $x\in\mathcal{X}$, the label is usually predicted as $\hat{Y}(x)=\arg\max_{k}\hat{P}_k(x)$.  While the performance of prediction can be evaluated by various metrics,  we consider the widely-used
top-1 accuracy, which
can be formally expressed as $acc= \mathbb{E}\left[ \mathbf{1}(Y=\hat{Y}(X))\right]$.
 In practice, however, a test dataset $\mathcal{D}^{\mathrm{T}}$ is usually used to
 empirically evaluate the accuracy, which is called the test accuracy:
\begin{equation}
\hat{acc}=\frac{1}{|\mathcal{D}^{\mathrm{T}}|} \sum_{(x,y)\in \mathcal{D}^{\mathrm{T}}} \mathbf{1}(y=\hat{Y}(x)).
\end{equation}

\subsection{Results on MNIST}
We begin our experiment of semi-supervised learning on a simple dataset --- MNIST,
which includes 60000 handwritten digit images with labels for training and 10000 images with labels for testing \cite{MNIST}. We use only 100 images (including 10 images for each digit) and their corresponding labels from the training dataset as labeled samples, and use the rest of images in the training dataset as unlabeled samples.
The test accuracy is calculated on the entire testing dataset of 10000 images.

\begin{table}[t]
	\setlength{\tabcolsep}{3pt}
	\centering
	\begin{tabular}{ p{2.4cm} | p{1.3cm}| p{2.4cm}| p{1.3cm} }
		\hline
		\textbf{Knowledge (Accurate)} & \textbf{Test Accuracy }&	\textbf{Knowledge (Imperfect C-II)} & \textbf{Test Accuracy}\\
		\hline
		\textbf{No Knowledge} & 0.8372& \textbf{No Knowledge} & 0.8372 \\
		\hline
		\textbf{One Hot} & 0.8725 &	\textbf{One Hot} & 0.8725 \\
		\hline
		\textbf{C-I} & 0.9224&	\textbf{C-I} & 0.9224 \\
		\hline
		\textbf{C-II} & 0.9654 & 	\textbf{C-II} & 0.9425\\	
		\hline
		\textbf{One Hot \& C-I} & 0.9396 &	\textbf{One Hot \& C-I}&0.9396 \\
		\hline
		\textbf{One Hot \& C-II} & 0.9686 &\textbf{One Hot \& C-II} &0.9590\\
		\hline
		\textbf{C-I \& C-II} &0.9733 &\textbf{C-I \& C-II} &0.9678 \\
		\hline
		 \textbf{One Hot \& C-I \& C-II} & 0.9775 &\textbf{One Hot \& C-I \& C-II} & 0.9687\\
		\hline
	\end{tabular}%
	\caption{Test accuracy given different combinations of domain knowledge on MNIST. (\textbf{C-I}: \textit{Constraint}-I;  \textbf{C-II}: \textit{Constraint}-II). }
	\vspace{-0.4cm}
	\label{table:mnist}
	
\end{table}

\begin{figure*}[!t]	
	\centering
	\subfigure[MNIST]{
			\label{fig:contribution_mnist}	\includegraphics[width=0.32\linewidth]{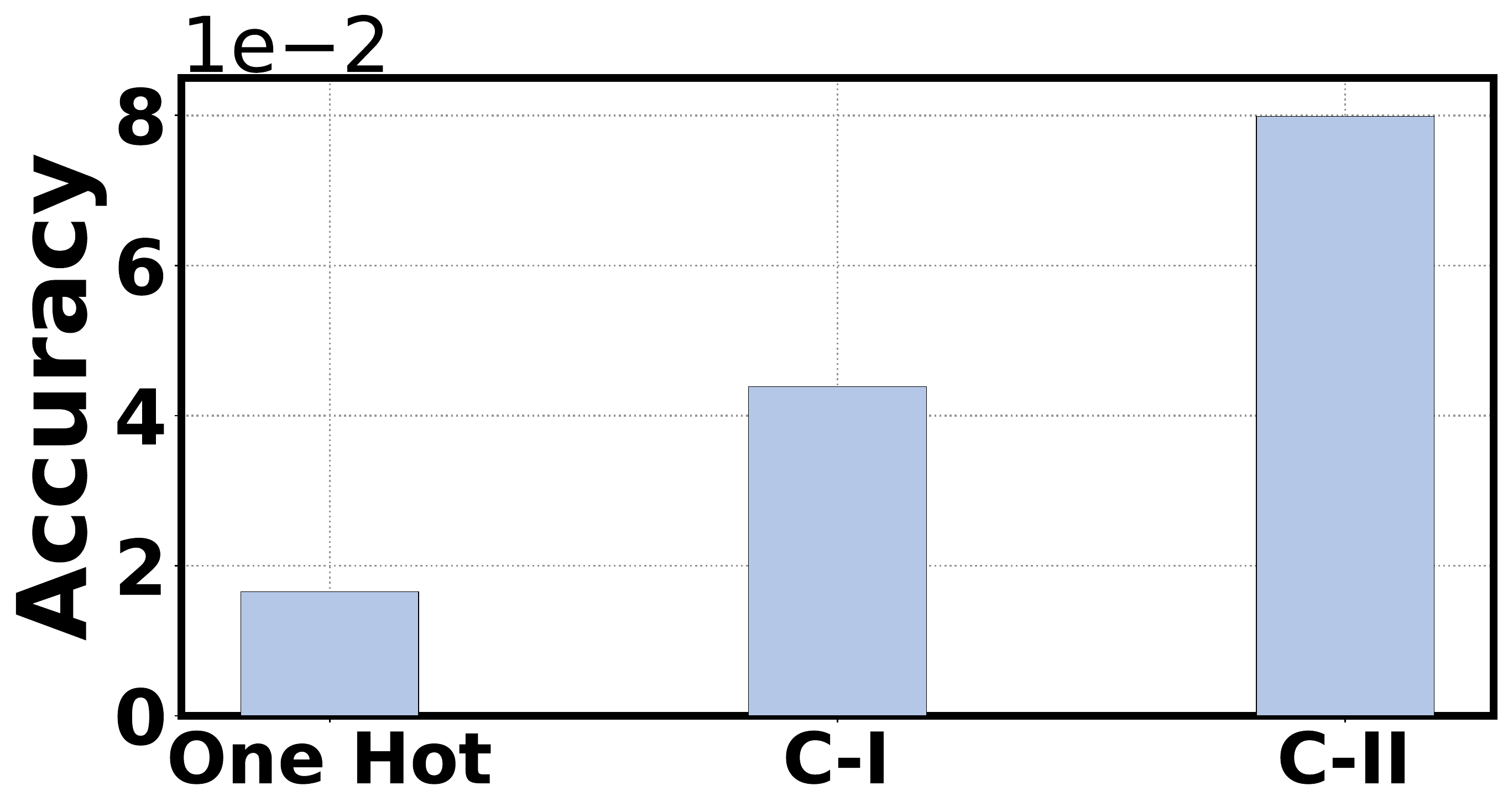}
	}
	\subfigure[MNIST (Imperfect C-II)]{
			\label{fig:contribution_mnist_corrupt}
		\includegraphics[width=0.32\linewidth]{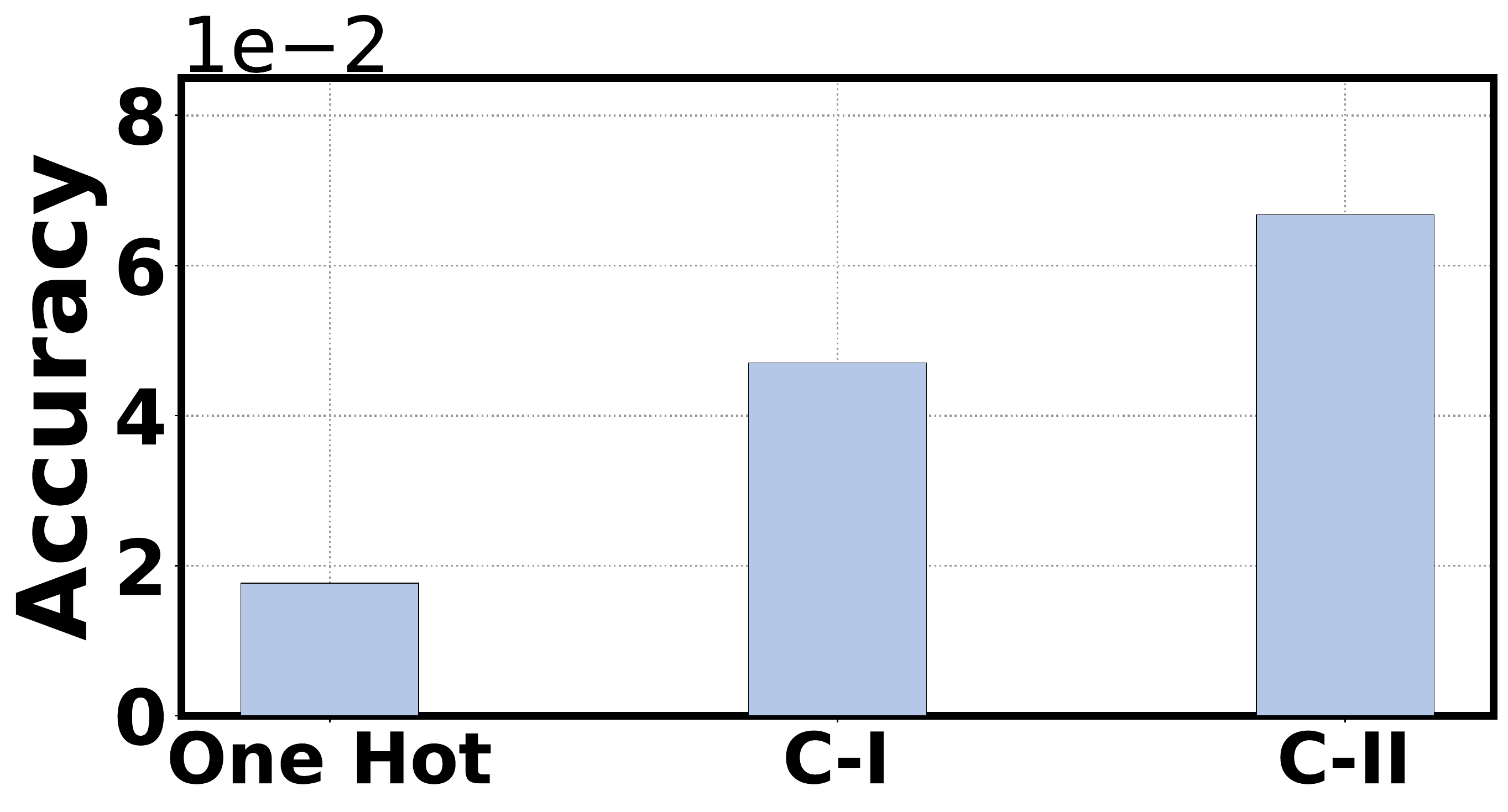}
	}
	\subfigure[CIFAR10]{
		\label{fig:contribution_cifar10}
		\includegraphics[width=0.32\linewidth]{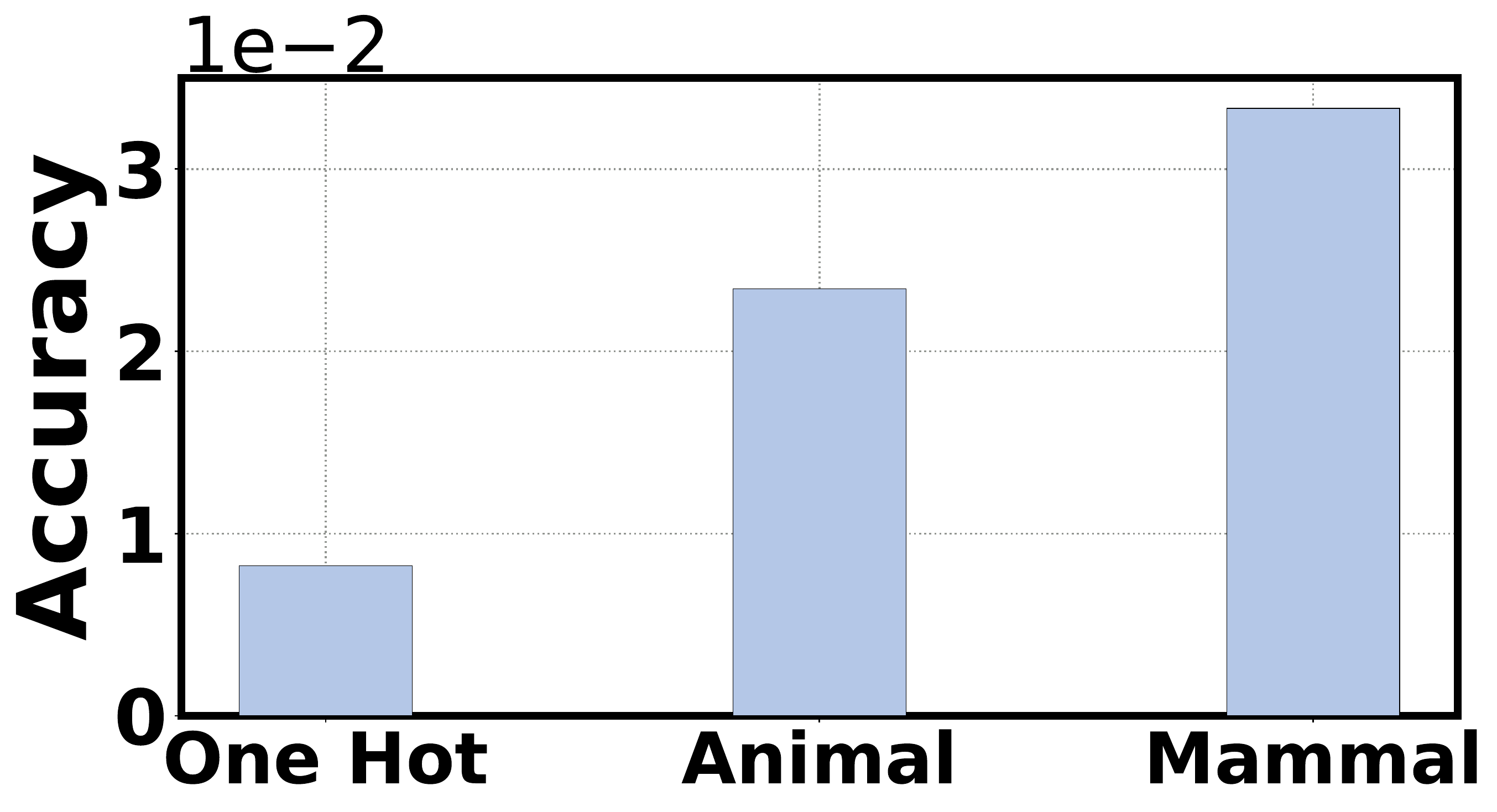}
	}
	\centering
	\caption{Domain knowledge Shapley in terms of test accuracy improvement
under the setting of semi-supervised learning.
	}

\end{figure*}

We consider a domain knowledge set \{\textit{One Hot}, \textit{Constraint}-I, \textit{Constraint}-II\}. The knowledge \textit{One Hot} as considered in \cite{Semantic_Loss_smbolic_xu2018semantic} means that each input has only one correct label, whose knowledge-based loss is expressed in Eqn.~\eqref{loss_onehot}. The knowledge \textit{Constraint}-I specifies whether or not the digit of an image belongs to the interval $[0,3]$, while the knowledge \textit{Constraint}-II informs us of whether or not the digit of an image is in the interval $[0,6]$.
Additionally, we quantify the knowledge values under the setting where  \textit{Constraint}-II in the knowledge set is imperfect. Specifically, 
we consider that 10\% of the unlabelled samples have wrong information about  \textit{Constraint}-II:
if the true label of a sample with corrupted
knowledge \textit{Constraint}-II is in the interval
$[0,6]$,
the corrupted knowledge \textit{Constraint}-II believes it is out of the interval $[0,6]$ and vice versa.
The loss based on knowledge \textit{Constraint}-I and \textit{Constraint}-II can both be calculated according to Eqn.~\eqref{loss_know}.

Our training setting follows  \cite{Semantic_Loss_smbolic_xu2018semantic}.
Specifically, we consider a DNN with four fully-connected hidden layers having 1000, 500, 250, 250 hidden neurons respectively, a batch normalization layer and a dropout layer with dropout probability 0.5. The weight to balance the cross-entropy loss and the knowledge-based loss is 0.1. During training, each batch has 16 images with corresponding labels and 16 images without labels (100 labeled images are reused to balance the number of labeled and unlabeled examples). Adam optimizer with learning rate $10^{-4}$ is used for updating weights followed by a tuning phase with learning rates $5\times10^{-5}, 10^{-5}, 5\times10^{-6}$ and $10^{-6}$. Note that since we only have three different pieces of domain knowledge
and our DNN has a reasonably low training cost, we re-train the DNN given
each combination of domain knowledge without Monte-Carlo sampling.

The test accuracies given each combination of domain knowledge  are shown in Table~\ref{table:mnist}. Based on the accuracy results, we calculate the value of each
domain knowledge in Fig.~\ref{fig:contribution_mnist}.
We see that the knowledge \textit{Constraint}-II has the largest contribution,
followed by the knowledge \textit{Constraint}-I, in terms of the test accuracy.
The knowledge \textit{One Hot} contributes the least, since it is simply
common knowledge. Our study provides a rigours quantification of values
of different knowledge.

In addition, we show in Table~\ref{table:mnist} the test accuracies given different domain knowledge combinations under the setting
where \textit{Constraint}-II is imperfect.
Accordingly, the quantified contribution of knowledge
in terms of test accuracy is shown in
Fig.~\ref{fig:contribution_mnist_corrupt}.
 We can see in Fig.~\ref{fig:contribution_mnist} that the contribution of imperfect knowledge \textit{Constraint}-II becomes naturally less than that of accurate knowledge \textit{Constraint}-II, while the contribution  of knowledge \textit{One Hot}  and \textit{Constraint}-I increases slightly. This implies that the performance degradation
is mostly caused by the imperfect knowledge \textit{Constraint}-II, which is consistent with our intuition. Interestingly, although the knowledge
\textit{Constraint}-I is more accurate than \textit{Constraint}-II, it does not provide
as much contribution to the test accuracy improvement as \textit{Constraint}-II.

\subsection{Results on CIFAR10}
\begin{table}[t]
	\setlength{\tabcolsep}{3pt}
	\centering
	\begin{tabular}{l|c}
		\hline
		\textbf{Knowledge Combinations} & \textbf{Test Accuracy} \\
		\hline
		\textbf{No Knowledge} & 0.7558 \\
		\textbf{One Hot} & 0.7716 \\
		\textbf{Animal} & 0.7948 \\
		\textbf{Mammal} & 0.8024\\
		\textbf{One Hot \& Animal} & 0.7983 \\
		\textbf{One Hot \& Mammal} & 0.8105 \\
		\textbf{Animal \& Mammal} & 0.8177 \\
		\textbf{One Hot \& Animal \& Mammal} & 0.8208\\
		\hline
	\end{tabular}%
	\caption{Test accuracy given different combinations of domain knowledge on CIFAR10.}
	\label{table:cifar10}
	\vspace{-0.5cm}
\end{table}

We also run experiments on the CIFAR10 dataset, which includes images with $32\times32$ pixels and 10 classes (airplane, automobile, bird, cat, deer, dog, frog, horse, ship, truck) in the label space \cite{CIFAR}. It has 50000 training samples and 10000 test examples. We use only 10\% of the training images with labels (5000 labeled images), and another 15000 training images without labels for training. For testing, we use all the 10000 test images in the test dataset.

Our domain knowledge set has three different types of symbolic knowledge \{\textit{One Hot, Animal, Mammal}\}. The knowledge \textit{One Hot} provides the information that each image has only one correct label, whose knowledge-based loss is expressed in Eqn.~\eqref{loss_onehot}. The knowledge \textit{Animal} provides the information whether or not the label of an image belongs to animals. With the knowledge \textit{Animal} , we can decide if the label of an input is in the label subset $\left\lbrace\text{bird, cat, deer, dog, frog, horse}\right\rbrace$, even though the exact label is unknown. The knowledge \textit{Mammal} provides information of whether or not the label of a training image belongs to mammals. Similarly, the knowledge \textit{Mammal} specifies whether the label of a training image is in the label subset \{\text{cat, deer, dog, horse}\}. The loss based on knowledge \textit{Animal} and \textit{Mammal} can be calculated by Eqn.~\eqref{loss_know}.

In our experiment, Ladder Net \cite{Semi-supervised_learning_pitelis2014semi, Semantic_Loss_smbolic_xu2018semantic} is used as the neural architecture. The weight to balance the cross entropy loss and knowledge-based loss is set as 0.1.   A batch of 512 inputs consist of both labeled images and unlabeled images are used for each weight update. To balance the number of labeled and unlabeled inputs, the labeled inputs are reused. Training is performed by the momentum algorithm with a learning rate of 0.1, followed by a tuning phase with learning rates 0.05, 0.01 and 0.0001.

After training the neural network with different combinations of domain knowledge, we evaluate the test accuracy as shown in Table~\ref{table:cifar10}.
The domain knowledge Shapley in terms of the test accuracy improvement is calculated
and shown in Fig.~\ref{fig:contribution_cifar10}. We can see that the knowledge \textit{Mammal} is the most informative one and hence has the largest contribution to the test accuracy, while the common-sense knowledge \textit{One Hot} contributes the least to the test accuracy improvement. The knowledge \textit{Animal} has a contribution between the other two knowledge.

%% file: conclusion.tex
\section{Conclusion}
In this paper, we study valuation of domain knowledge and
 provide a first step towards
quantitative understanding of contribution of domain knowledge injected
into informed machine learning.
Specifically, we propose domain knowledge Shapley to quantify the values of domain knowledge in terms of the learning performance improvement for machine learning tasks in a fair manner. We run experiments on both MNIST and CIFAR10 datasets using semi-supervised learning and quantitatively measure the fair value
 of different domain knowledge.
 Our results 
 also quantify the extent to which imperfect domain knowledge can affect the learning performance (for the MNIST dataset).